# A Review of Intelligent Practices for Irrigation Prediction

Krupakar Hans
SSN College Of Engineering
Chennai, India
E-mail: hanskrupakar1994@gmail.com

Akshay Jayakumar
SSN College Of Engineering
Chennai, India
E-mail: akshay.jayakumar1995@gmail.com

Dhivya G
SSN College Of Engineering
Chennai, India
E-mail: dhivya2204@gmail.com

**ABSTRACT:**

**Population growth and increasing droughts are creating unprecedented strain on the continued availability of water resources. Since irrigation is a major consumer of fresh water, wastage of resources in this sector could have strong consequences. To address this issue, irrigation water management and prediction techniques need to be employed effectively and should be able to account for the variabilities present in the environment. The different techniques surveyed in this paper can be classified into two categories: computational and statistical. Computational methods deal with scientific correlations between physical parameters whereas statistical methods involve specific prediction algorithms that can be used to automate the process of irrigation water prediction. These algorithms interpret semantic relationships between the various parameters of temperature, pressure, evapotranspiration etc. and store them as numerical precomputed entities specific to the conditions and the area used as the data for the training corpus used to train it. We focus on reviewing the computational methods used to determine Evapotranspiration and its implications. We compare the efficiencies of different data mining and machine learning methods implemented in this area, such as Logistic Regression, Decision Tress Classifier, SysFor, Support Vector Machine(SVM), Fuzzy Logic techniques, Artifical Neural Networks(ANNs) and various hybrids of Genetic Algorithms (GA) applied to irrigation prediction. We also recommend a possible technique for the same based on its superior results in other such time series analysis tasks.**

**INTRODUCTION:**

Water scarcity is becoming a major issue throughout the world today, which in turn has increased the threat of a major food crisis. Therefore, there needs to be an efficient method to utilize the available resources judiciously. As agriculture is one of the largest water consuming sectors, managing irrigation levels can play an important part in saving water for other purposes. This is also due to the fact that 25% of the water taken for agriculture is wasted due to poor management[1,2]. In order to maximize water productivity and improve water management, application of statistical methods have become significant.

As far as India is concerned, agriculture is the largest contributor to the GDP. As of 2010-11, the agricultural sector has contributed 14.2 percent of the GDP. Also, India's agricultural exports account for 1.4% of the world trade in agriculture[3]. Therefore, there is an increasing need to develop a prediction system effective management and utilization of soil nutrients and water resources. These systems based on spatial database on agriculture can improve agricultural management in India[4,5].

The earliest methods to address this issue involved computational models. In these methods, the focus lay on correlation between physical factors alone. Since these models relied on empirical scientific equations, they are only suitable in perfectly ideal conditions. They do not take into account the uncertainties and variations from ideal conditions that were used in determination of their implications that are found in real time data[6–8].

Over the last 15 years, there have been several applications of various data mining and prediction models in this area that have made coming up with an irrigation management and prediction system possible. These techniques are powerful, explorative and can perform intensive analysis on the given data. Some of the techniques used include Logistic Regression, Decision Tree Classifiers, SysFor, Support Vector Machines, Fuzzy Logic Classifiers and Artifical Neural Networks(ANNs).

The models described for irrigation prediction system are generally developed and trained using a large amount of historical data (training data) of entities or features that would influence the amount of water that is required in exactitude or at least inside of a coherent range. Once built, the models are used on real time data not used in training i.e. to make a prediction of the amount of irrigation necessary to sustain the crop's healthy growth. The nature of these models would be such that even though the algorithm allows for generalization to randomized test data, the predictions are most correspondent specifically to the area and conditions used in the training data, thereby trying to account for the otherwise invisible variations specific to that land and surroundings[1,9–11].

The objective of this paper is to study and review the computational various data mining techniques in the field of hydrology and to compare the efficiencies of the predictions made in the same. Section 2 of this paper looks at all the parameters or sources of features used in order to facilitate prediction of irrigation water requirement of crops. Section 3 of this paper reviews the computational parameter evapotranspiration and its implications and the different data mining techniques and the data analysis areas used in the field so far. Section 4 compares the accuracy of prediction of the different techniques discussed in Section 3. Section 5 describes a novel RNN LSTM model that is proposed in this paper that can be used to obtain better results, a richer contextualization and better semantic correlations found in the parameters used in real-time data[12–17]. Section 6 concludes the paper.

**SOURCES INFLUENCING IRRIGATION DEMAND:**

The sources of parameters that affect the process of crop irrigation usually tend to vary according to the terrain that agriculture is practiced. While some generic ones apply to all kinds of lands, we are going to restrict the discussion about the sources of feature vectors to terrains at ground level[18]. This list of features is by no means exhaustive and there are more and more parameters that can be considered with the advent of the new age Internet of Things sensors that can track various things that were previously not possible to track[19,20]. The sources that make irrigation water demand prediction possible can be broadly categorized into meteorological factors, crop input and agricultural parameters.

1. Meteorological Factors

Meteorological factors are important in determining the water requirement in crop irrigation because there are areas of the world, like in India, where the primary source of water is rainfall. Combined with this, there are also other factors that have the potential to affect the amount of water required for the efficient management and yield of crops. One of the main factors in this respect is the temperature parameters: maximum temperature, minimum temperature and average temperature. The effect on the amount of sunshine that a region requires is undeniably crucial to the plant's water needs as too much heat would cause lesser water availability to crops in general[21].

The speed of winds also tends to affect the amount of irrigation required, albeit not as much as temperature does. It is generally a practice to use a metric like Crop Water Stress Index (CWSI) to determine how much water is required for a sustained and healthy growth of plants. CWSI is the relative transpiration rate occurring from a plant by means of the temperature of the plant and the measure of humidity of the air in terms of vapour pressure deficit. It has been found that during the calculation of CWSI, predictions made during conditions of low wind speed tend to be in excess of the actual values and vice versa[22].

Other factors affecting crop irrigation requirement are rainfall, solar radiation intensity and duration, precipitation etc. The effects of these factors on crop irrigation demand prediction are quite straightforward as these factors affect the surface water availability and to an extent, the amount of groundwater present directly[23,24].

## 2. Crop Input Factors

One of the most important criteria that determines the irrigation demand is the soil type that is used to grow the particular crops. Major types of soil used for various crops tend to have different effects as the water retention capacity of the soil heavily depends on the composition of the soil. The difference can be seen clearly when the water requirement of various crops that grow on various types of soil are compared together[25]. One other factor considered include the type of crop itself. The behavior of water retention and requirement of every crop varies as the water content of each and every part of the plant (shoots, flowers, leaves, fruits etc.) varies[26]. Another parameter that is important is the soil moisture stress, which is a metric used to determine the extent of water that is not present in the soil[27].

Also, there are several other crop data surface coefficients that change with time that influence the water requirement of plants. But the use of these parameters is something that is left up to the programmer because these parameters are found out using relationships of the other feature vectors used in the mining algorithm. It is generally not advisable to do so as precomputed values might push the model to be prone to the errors in these parameters that would occur[6,28–32].

## 3. Agricultural Factors

Some other obvious factors include the time series parameters that determine the amount required in a situation that is affected by the history of occurrence. One such parameter is the actual amount of water that is used in the time steps that have occurred before the one being trained for. Some other information like amount of time remaining till the harvest are also considered sometimes so as to enforce pressurized irrigation behavior that the mining algorithm can learn from the best patterns possible as well[33].

**DESCRIPTION OF METHODS**:

The methods described in the following sections include ones that have been used for irrigation water prediction across all the parts of the world[34]. These methods have been used in specific parts of the world where electronically collected data about the conditions and the parameters is occurring. In the techniques discussed below, Evapotranspiration is the only one that is a computational method. All of the other methods discussed fall into the category of data mining and prediction algorithms based on historical context. Some of the comparative analyses specified in this paper comes from the work by Khan et al. from the comparative study he performed on the various data mining algorithms in terms of their performance of prediction by using the Coleambally Irrigation Area (CIA) in Australia[35].

## 1. Evapotranspiration

One of the really popular computational methods used to semantically predict the amount of water required is called EvapoTranspiration. EvapoTranspiration (ET) is a semantic measure of the total water loss that occurs due to evaporation and transpiration which specifies the amount of water essential for growth of plants. It is expressed as a depth (usually in inches). ET is based on a number of factors that can include: local temperature, precipitation, cloud cover, solar radiation, and the type of plants you are growing. The rate expresses the amount of water lost from a cropped surface in units of water depth in a unit area of land taken in the same units as the depth. The time unit can be an hour, day, decade, month or even an entire growing period or year. The parameter is usually compared with a Reference EvapoTranspiration ($ET_0$) that is used to study the demand of water requirement independent of crop type and variations in agricultural practices used. $ET_0$ is used to compute the ET on a hypothetical grass reference crop with specific characteristics[6,36,37].

Since this method involves modeling a scientific equation, there is no learning phase and the parameters for the day are simply substituted into the equation in order to calculate the score. Since empirical equations are considered, this method performs poorly on real-time scenarios due to uncertainties that are not taken into account in these equations.

The Evapotranspiration measure is derived from the FAO Penman-Monteith equation is computed as:

$$ET_o = \frac{\Delta(R_n - G) + \rho_a C_p (\delta e) g_a}{(\Delta + \gamma(1 + g_a/g_s))L_v}$$

where $L_v$ is the volumetric latent heat of vaporization i.e. the energy required per water volume vaporized, E is the mass water evapotranspiration rate in g / sec m², ETo is the water volume evapotranspired in mm / sec, $\Delta$ is the rate of change of saturation specific humidity with air temperature in Pa / K, Rn is the net irradiance in W / m² which is the external source of energy flux, G is the ground heat flux which is a little difficult to measure, in W / m², $C_p$ is the Specific heat capacity of air in J / kg K, $\rho_a$ is the dry air density in kg / m³, $\delta e$ is the specific humidity in Pa $g_a$ is the atmospheric conductance in m / s, $g_s$ is surface conductance in m / s, $\gamma$ is the Psychrometric constant which is close to 66 Pa / K[38].

**2. Logistic Regression**

Logistic regression is a very effective binary classifier that can be used to try to predict the water requirement as a time series data. Logistic regression can be Simple or Multivariate. Simple logistic regression is used to predict binary values. In this case, given the amount of water to be used, the model can predict whether the specification is adequate or not. This is a naive approach and predicted with a rather low accuracy.

Multivariate Logistic Regression takes into account various features to provide decisions[39,40]. This can be seen as the simplest feed-forward neural network with one hidden layer and therefore one weight. In this model, each class specifies the range of water required for irrigation. However, this model does not provide crop-specific decisions and in no way maps the temporal dependencies[35].

Logistic regression is modeled as:

$$P(Y_i = 1|\overline{X}_i = X) = \frac{e^{(W.X+B)}}{1 + e^{(W.X+B)}}, |X| = n$$

where X is the set of feature vectors of size n, $Y_i$ is the predicted class label, W and B are the weights and biases respectively that would be learned by means of the gradient descent algorithm[41].

**3. Decision Tree Classifier**

A decision tree deduces logical relationships between the class labels in the form of a hierarchical tree. The patterns explored by this algorithm is known as logic rules. Therefore, from the classifying attributes, classifier attributes are deduced[42]. Decision trees are made of nodes and leaves where the nodes represent the classifying attributes and the leaves represent the classifier attributes. These logical rules are framed with the use of information gain based on the C4.5 algorithm[43].

The model built using Decision Tree Classifiers has been found to provide greater accuracy than logistic regression. As with the problem of logistic regression, it is difficult to provide crop-specific predictions and model the temporal dependencies using this model[44].

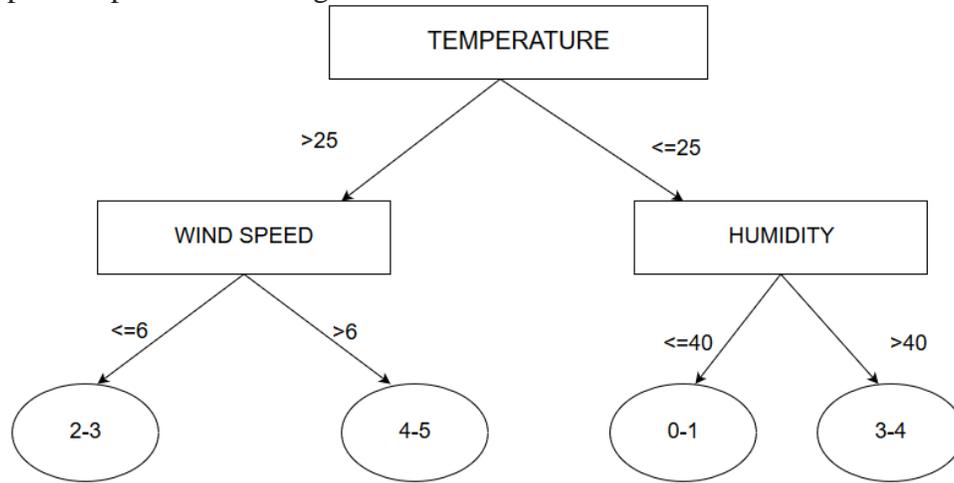

**Fig 1. Decision Tree Classifier**

**4.SysFor**

SysFor stands for Systematically developed Forests of multiple tress. This algorithm is a modification of the Decision Tree Classifier. The principle behind SysFor (Systematic Forests) is that better knowledge can be gained through extraction of better knowledge. This algorithm consists of several different steps. First, the 'goodness' of attributes and the split points in the dataset are determined using the gain ratio. The gain ratio is computed as:

$$Info(D) = - \sum_{i=1}^{m} p_i log_2(p_i)$$

$$Info_A(D) = \sum_{j=1}^{v} \frac{|D_j|}{|D|} I(D_j)$$

$$Gain(A) = Info(D) - Info_A(D)$$

where Info refers to the entropy and is computed as the probability of value being every unique element in the dataset in each and every feature. The gain ratio Gain is computed as the difference in entropy before the split and after the split occurred. This ratio is used to find the split points and the best features[45].

The next step is the specification of the number of trees used for the forest by the user. Each good attribute found is made the root attribute of a distinct tree. The first tree's root attribute is used to logically split the feature selection and prediction process horizontally and join the trees created in each split[46]. These steps are repeated till the user-defined number of trees are created in the forest. From this forest, the required classifier attributes are selected based on the logic rules of each tree[47,48].

The model used with this algorithm provides the best results with crop-specific predictions as found out in Khan et al[35]. This model has been found to be superior to even ANN models for the CIA land area, even though these models are not designed to be used in time series prediction.

## 5. Support Vector Machine(SVM)

An SVM tries to find the best amongst all the linear classifiers that can be possible between any 2 classes, over several high dimensional planes of data. This is achieved by construction of a hyperplane per dimension i.e. a set of hyperplanes in a higher dimensional space like the problem of crop irrigation demand prediction. The class labels used are identified in such a way that the distance between the hyperplanes used to find the best possible linear classifier is maximized. This ensures maximum differentiability of the class labels involved. The margin can be of two types- soft margin and hard margin. In soft margin, certain number of data points in each class are allowed to disobey the margins, which is not possible in hard margin SVMs. The margins are found using non-linear functions like sigmoid, tanh, gaussian function etc. Furthermore, if the number of features turns out to be very less, special Kernel functions can be used that form substituted combinations of the already existing features in a non linear space like sigmoid, tanh, exponential etc. A common practice used with these SVMs is the use of hierarchical microclustering to accommodate for huge datasets which all cannot be considered at once because of the complexities of computation of the centroid of a large number of data points. [49,50] The SVM classifier optimizes:

$$\max_W J(W) = \frac{2}{|W|}$$
$$S.T. \quad W.X + B = \begin{cases} \geq 1, y_i = 1 \\ \leq 1, y_i = -1 \end{cases} \quad \forall i \in [1, N]$$

where J is the cost function optimized based on the distance between the hyperplanes with respect to the linear classifier and $y_i$ is the output class label for item i.

The model developed with this algorithm has found to show surprisingly low accuracy. Also, the training period of this model is large. While this model does provide predictions that are crop-specific, the lack of ability to map time series correlations is also a problem.

## 6. Fuzzy Systems:

### A. Remote Sensing and Crop Models

Jones et al., in 2000, came up with an integrated module that combined remote sensing and crop models for decision support crop management. The model takes the remote sensed data, corrects and calibrates the data, and interprets this data, making it ready for training. Now, this data is sent to the crop and decision models for them to compute a fuzzy composite modeling scheme. The entire model is supplemented with ancillary data i.e.

various other dependent parameters like soil type, rainfall etc. The farm management decision system involved 3 main decision-making criteria: profitability, environment and sustainability. A simple fuzzy composite programming to try to estimate various parameters relevant to the agricultural domain. It is a farm management system on the whole more than a targeted water level predictor[51].

**B. Fuzzy Logic Crop Water Stress Index (FL-CWSI)**

Around the same time, Al-Faraj et al. built a rule-based Fuzzy Logic Crop Water Stress Index (FL-CWSI) which returns an index between 0 and 1 corresponding to the metric CWSI using fuzzy logic techniques. This index was usually a result of sigmoid, Gaussian, absolute difference or a variety of other membership functions used in fuzzy logic[52]. It eliminated the necessity of using various other factors like aerodynamic resistance etc. These models were found to perform better than the rule based models used for irrigation prediction[53].

**C. ANFIS**

By 2007, a new model was used for prediction of daily irrigation water demand using Adaptive Neuro-Fuzzy Inference Systems (ANFIS) technique. This first order Sugeno fuzzy model has 2 inputs and one output. The inputs are carried through 5 layers before an output prediction is reached. The first layer passes the input through a membership function that transforms input and convert it into a fuzzy set range. The second layer provides a product of the two membership function results on the two inputs. The third layer normalizes the influence of various rules by taking a ratio of the firing rate. The fourth layer introduces various parameters that will be used to predict accurate values The learning of these parameters usually takes place by means of a time series algorithm like ARMA, AR, ARIMA etc. The fifth layer computes a weighted average of all the input signals out of the 4th layer[54]. These models use the backpropagation algorithm through the 5 steps in reverse to facilitate the training process. They have a much better accuracy predicting irrigation requirement when compared to the Auto Regression and the ARMA models when Root Mean Squared (RMS) and Mean Absolute Percentage Error (MAPE)[55].

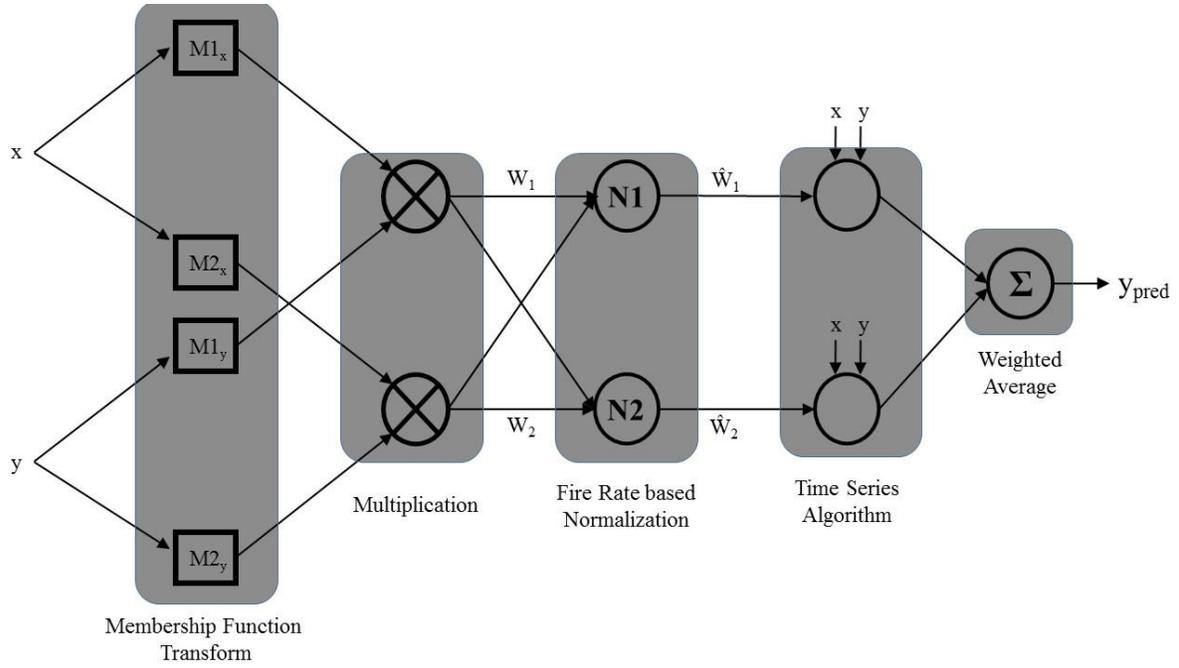

**Fig 2. ANFIS System with 2 inputs and 1 output**

**D. ARIMA**

In 2009, Wang et al. published a paper surveying the various kernel-based and other AI algorithms that were used for water flow prediction. Also, the analysis by Landeras et al. in 2009 revealed that the the ARIMA model and the ANN model, when used for calculation of evapotranspiration, gave a 6-8% reduction in the error percentage. Also, the work by Wang et al. showed that AR, ARMA, ANFIS models gave much better results when compared to the ANN models in the domain of water flow prediction[56]. ARIMA models are generalized class of models that can predict a stationary time series. Sometimes, stationarity detrending by means of taking the $d^{th}$ differential and modeling that as a stationary series, non-linear transformations etc. are carried out to ensure correct predictive results. A stationary series is constant in graph throughout the time series i.e. its central tendencies tend to remain constant inside of a range. This also acts as a disadvantage to this model as stationarity is a hard constraint to enforce on the dynamic real time data[57–59].

ARIMA(p,d,q) is given by:

$$X_t = \mu + \sum_{n=1}^{p} \phi_n X_{t-n} + \sum_{n=1}^{q} \gamma_n \varepsilon_{t-n} + \varepsilon_t$$

where $X_t$ is the predicted output at time step t, μ is the mean over the entire dataset, the summation from 1 to p is the autoregressive model that learns predictions based on a weighted sum of a predefined number of time step values, the summation from 1 to q is the moving average model that maps the error in prediction of every time step, $\xi_t$ is the error parameter that is permitted. The parameter d specifies the number of times the given input data needs to be differentiated in order to achieve a data distribution that is stationary, where the differentiation is carried out at each level by means of finding the difference between the timesteps. Of these parameters, $\xi_t$, $\Phi$ and $\gamma$ are the learnable parameters of the model[60].

**E. Fuzzy inference to single factor prediction**

By 2010, a new model that made use of fuzzy inference to single factor prediction method used to predict the amount of irrigation required. First, pre-recorded examples are taken and discretized into a universe U. Then a reasoning model of single factor prediction characterized by (e-f)-fold by e-degree compound fuzzy implicative propositional logic. The excessive dimensionality problem is taken care of using logic primitives that work to reduce the feature space. The inference rules present are taken and conformed to a required size set to ensure a rule based solution[61].

**F. Hybrid Model**

There have been a lot of hybrid models that have been used for time series prediction effectively that involve various combinations of ANNs, fuzzy logic and Genetic Algorithms in various other domains[61–65]. In 2010, Paulido-Calvo et al. came up with a hybrid model of combining CNNs with fuzzy membership function, optimized by a genetic optimization algorithm to test the efficacy of irrigation demand prediction. He found that the selection of the best out of 5 ANN models in triangular partitions and subjecting that through a fuzzy membership function, optimized by genetic optimization algorithms using IF-THEN rules gave the best results. The application of fuzzy algorithms was quickly followed by defuzzification algorithms to produce the required output. The genetic operations of reproduction, crossover and mutation were defined to facilitate the genetic optimization. This system beat the accuracy of the conventional ANN system[1].

**G. Inexact rough interval fuzzy linear programming**

Lu et al. tried to use inexact rough interval fuzzy linear programming method to generate conjunctive water allocation strategies. The rough interval method embodies dual-uncertain parameters quite robustly. The results from this model were comparable to the existing state-of-the-art even though the execution time of the algorithm was much better than the contemporary model whose performance is similar to this one. One of the main advantages in using this model is its ability to predict good results in both normal and special system conditions, making it a more generalized solution that can be achieved by selective targeting of features[66].

**H. Decision Support System**

In 2014, Mousa et al. developed a fuzzy based decision support system that evaluated evapotranspiration value using the computational formula before being used in training to predict the amount of irrigation required by means of use of the decision tree algorithm. The model took input from sensors of wind speed, temperature, humidity etc and also given as input were the irrigation strategies used and the equipment used to see if these factor into the prediction methodology. The irrigation strategy planned and equipment information gave the model good performance and considerably better speed of execution. This model gave accurate and fast results when compared to some of the other models like AR, ARMA etc.[67]

**7. ANNs**

Artificial Neural Networks are those architectures in which the input is morphed into a plane where a linear separation or regression is made possible by means of a non-linearity to warp the input into a plane where a straight line is enough. Various hyper-parameters used in an ANN would be hidden layer size, number of hidden layers, batch size, activation function, learning rate, optimization function, initialization strategy and normalization. The model is trained by means of various training algorithms like BackPropagation, radial basis function etc. The complexity of what the model can model is usually proportional to the number of layers in the architecture[68–70].

$$a(x) = \tanh(x)$$
$$y_{pred} = (a \circ a \cdots \circ a)_{n\ times}(W.X + B)$$

where a is the activation function used, which is usually any number of nonlinearities like sigmoid, tanh etc. The model is trained as a composition of activation functions that occurs the same number of times as the number of layers to give the final output. While these models operate in the nonlinear space and have much more scope for accommodating generic input, these models don't have the capabilities to map the time series dependencies found in irrigation demand prediction[71].

Around 2001, the performance of artificial neural network architectures was compared with most of the other models at the time like ARMA, AR, SVMs etc. It was found that ANNs performed consistently worse than the SMT and the ANFIS models[56]. But by 2003, ANNs were tweaked efficiently enough to become better in performance than the other models like ANFIS, ARMA etc.[72] Hardaha et al. did extensive work on tweaking the various parameters of the ANN to give an exploratory understanding on the dependence of each hyper-parameter to the accuracy of the model. It was found that training based on the radial basis function gave the best results for prediction of water requirement of wheat crops[73].

## A. Artificial Neuro-Genetic Networks

A hybrid ANN model that incorporates a genetic algorithm on top of an ANN to improve the accuracy of the model was proposed in 2009. This model used the Artificial Neuro-Genetic Networks (ANGN) to predict the irrigation requirement of the crops. Initially, the collection of ANNs of size N are taken to be the entire population. A first generation would contain an initialized collection of ANNs all with their own hyper-parameters. The output from the initial population is sent into the ANN chosen is then used to train the particular input. Now the accuracies of the different models are checked to see which ANGN selection would result in the best performance. To optimize the entire process, a pareto front is added before the output is specified. Objective evaluation functions F1 and F2 are first calculated over the entire corpus and are further used in the ANGN selection process[74].

A study by Khan et al. in 2013 compares different AI models and their accuracies when it comes to irrigation prediction. It was found that of all the models, the model with the 3-fold cross validation multiple decision trees SysFor model gave the best overall results. But the actual water content required by the crop was accurately predicted by ANNs really accurately. The difference in error percentage between ANNs and SysFor was almost 20%. Thus, it was concluded that SysFor, ANN and decision tree techniques are the most suitable for the task of irrigation prediction[35].

## COMPARISION OF PERFORMANCES

| Model Used | Dataset / Study area | Time Step | Accuracy (%) | Reference |
|---|---|---|---|---|
| Logistic Regression | Coleambally, Australia | 1 day | 56 | Khan et al.[35] |
| Decision Tree Classifier | Coleambally, Australia | 1 day | 74 | Khan et al.[35] |
| SysFor | Coleambally, Australia | 1 day | 78 | Khan et al.[35] |
| Support Vector Machines | Coleambally, Australia | 1 day | 64 | Khan et al.[35] |
| Remote Sensing and Crop Models | MAC, University of Arizona | Variable (Rule Based) | 58 to 78 | Jones et al.[51] |
| FL-CWSI | Horticulture Dept., UNL | Index – no time series info | CWSI Comparison Only | Al-Faraj et al.[53] |
| ANFIS | Chania in Crete, Greece | 1 day | MAPE comparison Only | Atsalakis et al.[55] |
| ARIMA | Álava, Basque County, Spain | 1 week | ET Comparison Only | Landeras et al.[59] |
| Fuzzy inference to Single Factor Prediction | Shangqiu, Henan, China | 1 year | Value Comparison Only | Chen et al.[61] |
| Hybrid ANN, Fuzzy, GA | Cordoba province, Spain | 1 day | 79.73 (in terms of 20.27 SEP %) | Pulido-Calvo et al.[1] |
| Inexact rough interval fuzzy | Custom dataset | Custom | Comparison Only | Lu et al.[66] |
| Decision Support System | Baghdad, Iraq (Simulation) | 1 month | ET Comparison Only | Mousa et al.[67] |
| ANNs | Bhakra Canal system, Rajasthan, India | 1 month | $r^2$ Comparison Only | Hardaha et al.[73] |
| ANGN | Bembézar Irrigation District (Spain) | 1 day | 87.37 (in terms of 12.63 SEP %) | Perea et al.[74] |

**Table 1 Comparison of Models**

Simply using fuzzy system architectures will help capture the randomness in the representation and the fading of hard boundaries at inflection points, but will not do much about the sequential nature of the data. ARIMA models take into context the temporal dependencies but only work for univariate regression. Multivariate regression is essential for irrigation prediction. Even the ANFIS architecture is not designed to capture time series dependencies. ANNs are very good at prediction based on training, but lack the capability of inferring semantic meaning from the sequential flow. SVMs and other kernel based methods are designed to be trained on normal data and will not be able to capture the essential sequential information. ANGN also fails to map temporal dependencies when reading features and predicting outputs.

**PROPOSED MODEL**

The above mentioned fallacies have left us in an unfortunate situation of choosing amongst a pool of ordinary choices and simply settling for whatever accuracy we can obtain. Hence, we propose a novel methodology of using a sequence learning based recurrent neural network (RNN) model that uses the LSTM activation function to model for irrigation requirement so that it doesn't have memory problems on long input streams. The input historical context will be stored inside the context vector which will be passed on through the mapping of the output through time in the network[75–78].

$$a(x) = tanh(x)])$$
$$h_{t=a(W_h h_{t-1} + W X_t + B)}$$

where a is the activation function, $X_t$ is the input at time step t, W, $W_h$ and B are learnable parameters of the model and h refers to the hidden layer output that occurs during every time step t.

The underlying architecture of the should contain three to five hidden LSTM or GRU layers, of size n, that are all back-propagated during training. The model uses BackPropagation Through Time (BPTT) algorithm for training and tweaking of the layer weights. The model will typically contain thousands of weight vectors all trained for each sequence input. The input will be in the form of features like temperature, rainfall, humidity etc. determined using real time sensors. A fixed context needs to be assigned for how much of the historic context is required for prediction. Every time step will involve processing all the features of that particular time step[13,79,80].

A typical LSTM circuit is described by:

$$i_t = sigm(W_i[h_{t-1}, X_t] + B_i)$$
$$f_t = sigm(W_f[h_{t-1}, X_t] + B_f)$$
$$o_t = sigm(W_o[h_{t-1}, X_t] + B_o)$$
$$c_t = f_t c_{t-1} + tanh(W_c[h_{t-1}, X_t] + B_c)$$
$$h_t = o_t \tanh(c_t)$$

where i is the input gate, f the forget gate, o the output gate, c the cell gate and h the hidden layer output. The interactions in the cell state by means of which the mapping of the backpropagated gradients occurs not separately but alongwith a cell state interaction that ensures that the gradient never reaches zero[81].

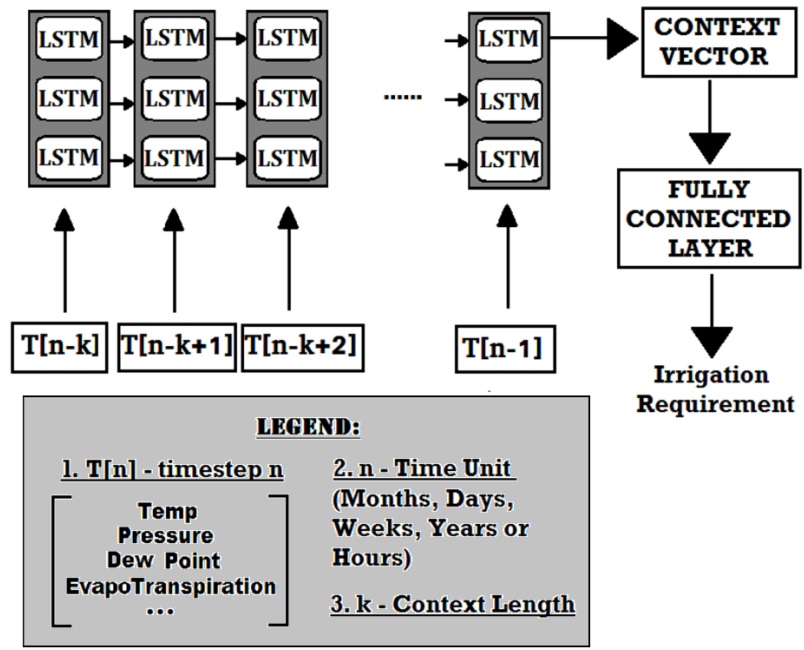

**Fig 3. RNN LSTM Model Proposed**

The advantages of using such an architecture are plentiful. The RNN is robust enough to map the intrinsic variations in observations that occur due to various factors like global warming, faulty equipment etc. Uncertainty will be automatically detected and ignored by the RNN. Also, RNNs are designed to incorporate sequence information inside of the hidden layer vector as context. The downsides to this model are the computational feasibility and the availability of hardware (GPUs) powerful enough to train a big model in less time, and the requirement of a huge corpus for training, which is why it was not possible to implement this model. Nevertheless, the runtime of the model will be really fast because testing is merely a few multiplications. Another advantage of this model is its efficiency in storage. Deep learning models only occupy very small amounts of space (in hundreds of MB) when compared to the traditional AI algorithms (a few GB)[77].

Because of all these factors, and the recent success of RNN sequence to sequence architectures, the application of this architecture should improve the current SOTA by quite a bit.